\documentclass[a4paper,twoside]{article}

\usepackage{amssymb}
\usepackage{amstext}
\usepackage{amsmath}
\usepackage{multicol}
\usepackage{pslatex}
\usepackage{apalike}
\usepackage{fancyhdr}
\usepackage{scitepress}
\usepackage[small]{caption}
\usepackage{graphicx}
\usepackage{algorithm}
\usepackage[noend]{algorithmic}
\usepackage{qtree}
\usepackage{color}
\usepackage[utf8]{inputenc} 
\usepackage[table]{xcolor}

\begin{document}
\onecolumn \maketitle \normalsize \vfill

\section{\uppercase{Introduction}}
\noindent The World Wide Web today contains an exterminated amount of information, mostly unstructured, under the form of Web pages, but also documents of various nature. 
During last years big efforts have been conducted to develop techniques of information extraction on top of the Web. 
Approaches adopted spread in several fields of Mathematics and Computer Science, including, for example, logic-programming and machine learning. 
Several projects, initially developed in academic settings, evolved in commercial products, and it is possible to identify different methodologies to face the problem of Web data extraction. 
A widely adopted approach is to define \emph{Web wrappers}, procedures relying on analyzing the structure of HTML Web pages (i.e. DOM tree) to extract required information.
Wrappers can be defined in several ways, e.g. most advanced tools let users to design them in a visual way, for example selecting elements of interest in a Web page and defining rules for their extraction and validation, semi-automatically; regardless of their generation process, wrappers intrinsically refer to the HTML structure of the Web page at the time of their creation. 
Thus, introducing not negligible problems of robustness, wrappers could fail in their tasks of data extraction if the underlying structure of the Web page changes, also slightly.
Moreover, it could happens that the process of extraction does not fail but extracted data are corrupted. 

All these aspects clarify the following scenarios: during their definition, wrappers should be as much elastic as possible, in order to intrinsically handle minor modifications on the structure of Web pages (this kind of small local changes are much more frequent than heavy modifications); although elastic wrappers could efficiently react to minor changes, maintenance is required for the whole wrapper life-cycle. 
Wrapper maintenance is expensive because it requires highly qualified personnel, specialized in defining wrappers, to spend their time in rebuilding or fixing wrappers whenever they stop working properly.
For improving this aspect, several commercial tools include notification features, reporting warnings or errors during wrappers execution. 
Moreover, to increase their reliability, data extracted by wrappers could be subject to validation processes, and also data cleaning is a fundamental step; some tools provide caching services to store the last working copy of Web pages involved in data extraction processes.
Sometimes, it is even more convenient to rewrite \emph{ex novo} a wrapper, instead of trying to find causes of malfunctioning and fixing them, because debugging wrapper executions can be not trivial. 
The unpredictability of what changes will occur in a specific Web page and, consequently, the impossibility to establish when a wrappers will stop working properly, requires a smart approach to wrapper maintenance.

Our purpose in this paper is to describe the realization and to investigate performances of an automatic process of wrapper adaptation to structural modifications of Web pages. 
We designed and implemented a system relying on the possibility of storing, during the wrapper definition step, a \emph{snapshot} of the DOM tree of the original Web page, namely a \emph{tree-gram}.
If, during the wrapper execution, problems occur, this sample is compared to the new DOM structure, finding similarities on trees and sub-trees, to automatically try adaptating the wrapper with a custom degree of accuracy.
Briefly, the paper is structured as follows: Section 2 focuses on related work, covering the literature about wrapper generation and adaptation.
In Section 3 we explain some concepts related to the tree similarity algorithm implemented, to prove the correctness of our approach. Section 4 shows details about our implementation of the automatic wrapper adaptation. Most important results, obtained by our experimentation, are reported in Section 5. Finally, Section 6 concludes providing some remarks for future work.

\section{\uppercase{Related Work}}
\noindent The concept of analyzing similarities between trees, widely adopted in this work, was introduced by Tai \cite{Tai1979}; he defined the notion of \emph{distance} between two trees as the measure of the dissimilarity between them. 
The problem of transforming trees into other similar trees, namely \emph{tree edit distance}, can be solved applying elementary transformations to nodes, step-by-step.
The minimum cost for this operation represents the tree edit distance between the two trees.
This technique shows high computational requirements and complex implementations \cite{Bille2005}, and do not represents the optimal solution to our problem of finding similarities between two trees. 
The \emph{simple tree matching} technique \cite{StanleyM.Selkow1977} represents a turning point: it is a light-weight recursive top-down algorithm which evaluates position of nodes to measure the degree of isomorphism between two trees, analyzing and comparing their sub-trees. 
Several improvements to this technique have been suggested: Ferrara and Baumgartner \cite{Ferrara2010}, extending the concept of weights introduced by Yang \cite{Yang1991}, developed a variant of this algorithm with the capability of discovering clusters of similar sub-trees.
An interesting evaluation of the simple tree matching and its weighed version, brought by Kim et al. \cite{Kim2007}, was performed exploiting these two algorithms for extracting information from HTML Web pages; we found their achievements very useful to develop automatically adaptable wrappers.

Web data extraction and adaptation rely especially on algorithms working with DOM trees. 
Related work, in particular regarding Web wrappers and their maintenance, is intricate: Laender et al. \cite{Laender2002} presented a taxonomy of wrapper generation methodologies, while Ferrara et al. \cite{Baumgartner2010} discussed a comprehensive survey about techniques and fields of application of Web data extraction and adaptation.
Some novel wrapper adaptation techniques have been introduced during last years: a valid hybrid approach, mixing logic-based and grammar rules, has been presented by Chidlovskii \cite{Chidlovskii2001}.
Also machine-learning techniques have been investigated, e.g. Lerman et al. \cite{Lerman2003} exploited their know-how in this field to develop a system for wrapper verification and re-induction.
Meng et al. \cite{20} developed the SG-WRAM (Schema-Guided WRApper Maintenance), for wrapper maintenance, starting from the observation that, changes in Web pages, even substantial, always preserve syntactic features (i.e. syntactic characteristics of data items like data patterns, string lengths, etc.), hyperlinks and annotations (e.g. descriptive information representing the semantic meaning of a piece of information in its context).
This system has been implemented in their Web data extraction platform: wrappers are defined providing both HTML Web pages and their XML schemes, describing a mappings between them. 
When the system executes the wrapper, data are extracted under the XML format reflecting the previously specified XML Schema; the wrapper maintainer verifies any issue and, eventually, provides protocols for the automatic adaptation of the problematic wrapper. 
The XML Schema is a DTD (Document Type Definition) while the HTML Web page is represented by its DOM tree.
The framework described by Wong and Lam \cite{Wong} performs the adaptation of wrappers previously learned, applying them to Web pages never seen before; they assert that this platform is also able to discover, eventually, new attributes in the Web page, using a probabilistic approach, exploiting the extraction knowledge acquired through previous wrapping tasks.
Also Raposo et al. \cite{Raposo2005} suggested the possibility of exploiting previously acquired information, e.g. results of queries, to ensure a reliable degree of accuracy during the wrapper adaptation process. 
Concluding, Kowalkiewicz et al. \cite{Kowalkiewicz2006} investigate the possibility of increasing the robustness of wrappers based on the identification of HTML elements, inside Web pages, through their XPath, adopting relative XPath, instead of absolute ones.

\section{\uppercase{Matching Up HTML Trees}}

\noindent Our idea of automatic adaptation of wrappers can be explained as follows: first of all, outlining how to extract information from Web pages (i.e. in our case, how a Web wrapper works); then, describing how it is possible to recover information previously extracted from a different Web page (i.e. how to compare structural information between the two versions of the Web page, finding similarities); finally, defining how to automatize this process (i.e. how to build reliable, robust automatically adaptable wrappers). 

Our solution has been implemented in a commercial product \footnote{Lixto Suite, www.lixto.com}; Baumgartner et al. \cite{baumgartner2009scalable} described details about its design.  
This platform provides tools to design Web wrappers in a visual way, selecting elements to be extracted from Web pages. 
During the wrapper execution, selected elements, identified through their XPath(s) in the DOM tree of the Web page, are automatically extracted.
Although the wrapper design process lets users to define several restricting or generalizing conditions to build wrappers as much elastic as possible, wrappers are strictly interconnected with the structure of the Web page on top of they are built. 
Usually, also slight modifications to this structure could alter the wrapper execution or corrupt extracted data. 

In this section we discuss some theoretical foundations on which our solution relies; in details, we show an efficient algorithm to find similar elements within different Web pages.

\subsection{Methodology}
\noindent A simple measure of similarity between two trees, once defined their comparable elements, can be established applying the \emph{simple tree matching} algorithm \cite{StanleyM.Selkow1977}, introduced in Section 2. 
We define \emph{comparable elements} among HTML Web pages, nodes, representing HTML elements (or, otherwise, free text) identified by tags, belonging to the DOM tree of these pages. 
Similarly, we intend for \emph{comparable attributes} all the attributes, both generic (e.g. \emph{class}, \emph{id}, etc.) and type-specific (e.g. \emph{href} for anchor texts, \emph{src} for images, etc.), shown by HTML elements; it is possible to exploit these properties to introduce additional comparisons to refine the similarity measure.
Several implementations of the \emph{simple tree matching} have been proposed; our solution exploits an improved version, namely \emph{clustered tree matching} \cite{Baumgartner2010}, designed to match up HTML trees, identifying clusters of sub-trees with similar structures, satisfying a custom degree of accuracy.

\subsection{Tree Matching Algorithms}
\noindent Previous studies proved the effectiveness of the \emph{simple tree matching} algorithm applied to Web data extraction tasks \cite{Kim2007,19}; it measures the similarity degree between two HTML trees, producing the maximum matching through dynamic programming, ensuring an acceptable compromise between precision and recall. 

As improvement to this algorithm, this is a possible implementation of \emph{clustered tree matching}: let \emph{d(n)} to be the degree of a node \emph{n} (i.e. the number of first-level children); let T(i) to be the i-\emph{th} sub-tree of the tree rooted at node T; let $t(n)$ to be the number of total siblings of a node \emph{n} including itself.

\begin{algorithm}
\caption{ClusteredTreeMatching($T^{'}$, $T^{''}$)}
\label{alg1}
\begin{algorithmic}[1]
    \IF{$T^{'}$ has the same label of $T^{''}$}
        \STATE $m \leftarrow$ $d(T^{'})$
        \STATE $n \leftarrow$ $d(T^{''})$
        \FOR{$i = 0$ to $m$}
            \STATE $M[i][0] \leftarrow 0$;
        \ENDFOR
        \FOR{$j = 0$ to $n$}
            \STATE $M[0][j] \leftarrow 0$;
        \ENDFOR
        \FORALL{$i$ such that $1\leq i\leq m$}
            \FORALL{$j$ such that $1\leq j \leq n$}
                \STATE $M[i][j] \leftarrow$ Max($M[i][j-1]$, $M[i-1][j]$, $M[i-1][j-1] + W[i][j]$) where $W[i][j]$ = ClusteredTreeMatching($T^{'}(i-1)$, $T^{''}(j-1)$)
            \ENDFOR
        \ENDFOR
        
        \IF{$m > 0$ AND $n > 0$}
        	\STATE return M[m][n] * 1 / Max($t(T^{'})$, $t(T^{''})$)
    		\ELSE
        	\STATE return M[m][n] + 1 / Max($t(T^{'})$, $t(T^{''})$)
    		\ENDIF
    \ELSE
        \STATE return 0
    \ENDIF
\end{algorithmic}
\end{algorithm}

\begin{figure*}
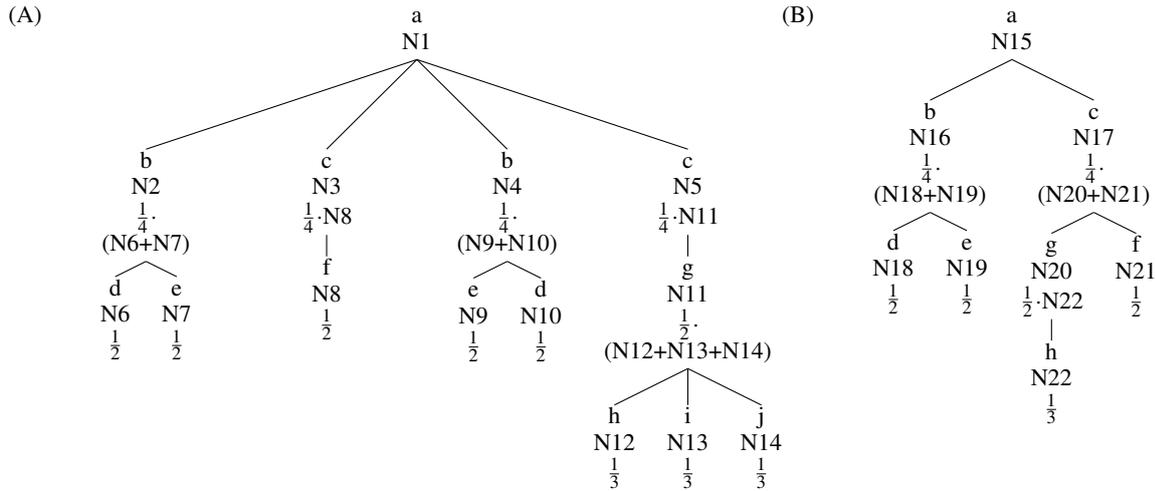


	\small (A)
	\Tree [.{a\\\small N1} [.{b\\\small N2\\$\frac{1}{4}\cdot$\\\small (N6+N7)} [.{d\\\small N6\\ $\frac{1}{2}$} ] [.{e\\\small N7\\ $\frac{1}{2}$} ] ] [.{c\\\small N3\\$\frac{1}{4}\cdot$\small N8} [.{f\\\small N8\\ $\frac{1}{2}$} ]  ] [.{b\\\small N4\\$\frac{1}{4}\cdot$\\\small (N9+N10)} [.{e\\\small N9\\ $\frac{1}{2}$} ] [.{d\\\small N10\\ $\frac{1}{2}$} ] ] [.{c\\\small N5\\$\frac{1}{4}\cdot$\small N11} [.{g\\\small N11\\$\frac{1}{2}\cdot$\\\small (N12+N13+N14)} [.{h\\\small N12\\ $\frac{1}{3}$} ] [.{i\\\small N13\\ $\frac{1}{3}$} ] [.{j\\\small N14\\ $\frac{1}{3}$} ] ] ] ]
	\small (B)
	\Tree [.{a\\\small N15} [.{b\\\small N16\\$\frac{1}{4}\cdot$\\\small (N18+N19)} [.{d\\\small N18\\ $\frac{1}{2}$} ] [.{e\\\small N19\\ $\frac{1}{2}$} ]  ] [.{c\\\small N17\\$\frac{1}{4}\cdot$\\\small (N20+N21)} [.{g\\\small N20\\$\frac{1}{2}\cdot$\small N22} [.{h\\\small N22\\ $\frac{1}{3}$} ] ] [.{f\\\small N21\\ $\frac{1}{2}$} ] ] ]

\caption{Two labeled trees, \emph{A} and \emph{B}, which show similarities in their structures.}
\label{fig1}
\end{figure*}

\noindent The main difference between the \emph{simple} and the \emph{clustered} tree matching is the way of assigning values to matching elements. The first, adopts a fixed matching value of 1; the latter, instead, computes some additional information, retrieved in the sub-trees of matched nodes. 

Omitting detail, provided in \cite{Baumgartner2010}, the \emph{clustered tree matching} algorithm assigns a weighted value equal to 1, divided by the greater number of siblings, computed between the two compared nodes (also considering themselves). 

Figure \ref{fig1} shows two similar simple rooted, labeled trees, and the way of assignment of weights that would be calculated by applying the \emph{clustered tree matching} between them. 

\begin{figure*}[!ht]%
\begin{center}
	\includegraphics[width=360px]{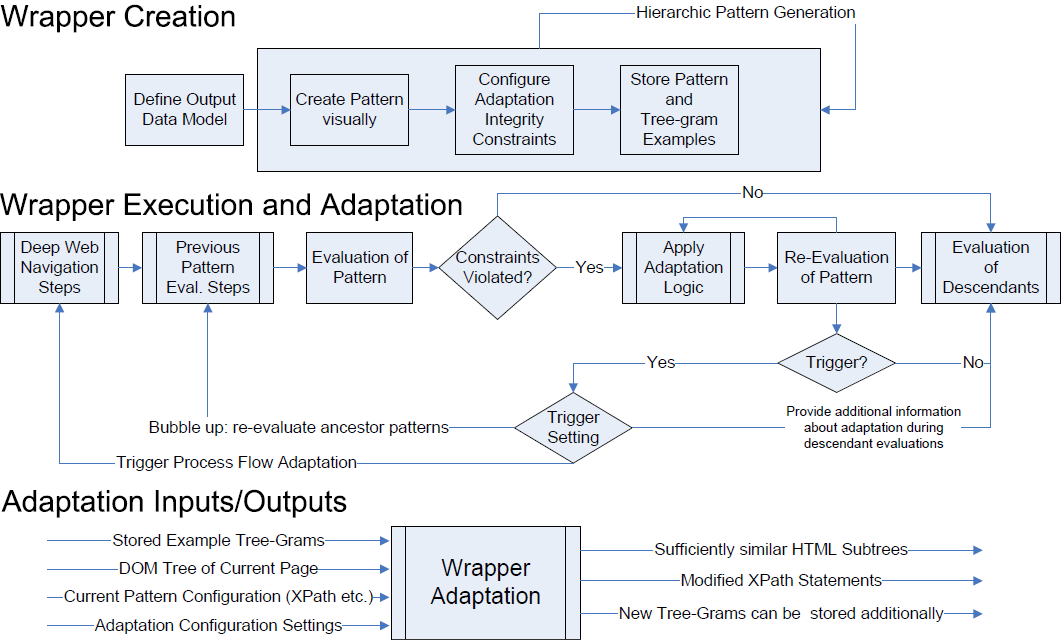}%
	\caption{State diagram of Web wrappers design and adaptation in the Lixto Visual Developer.}%
	\label{diagram}%
\end{center}
\end{figure*}

\subsubsection{Motivations}
\noindent Several motivations lead us to bring these improvements. 
For example, considering common characteristics shown by Web pages, provides some useful tips: usually, rich sub-levels (i.e. sub-levels with several nodes) represent list items, table rows, menu, etc., more frequently affected by modifications than other elements of Web pages; moreover, analyzing which kind of modifications usually affect Web pages suggests to assign less importance to slight changes happening in deep sub-levels of the DOM tree, this because these are commonly related to missing/added details to elements, etc.

On the one hand, \emph{simple tree matching} ignores these important aspects, on the other \emph{clustered tree matching} exploits information like position and number of mismatches to produce a more accurate result.

\subsubsection{Advantages and limitations}
\noindent The main advantage of our \emph{clustered tree matching} is its capability to calculate an absolute measure of similarity, while \emph{simple tree matching} produces the mapping value between two trees.
Moreover, the more the structure of considered trees is complex and similar, the more the measure of similarity established by this algorithm will be accurate. 
It fits particularly well to matching up HTML Web pages, this because they often own rich and variegated structures.

One important limitation of algorithms based on the tree matching is that they can not match permutations of nodes. 
Intuitively, this happens because of the dynamic programming technique used to face the problem with computational efficiency; both the algorithms execute recursive calls, scanning sub-trees in a sequential manner, so as to reduce the number of required iterations (e.g. in Figure \ref{fig1}, permutation of nodes [c,b] in \emph{A} with [b,c] in \emph{B} is not computed). 
It is possible to modify the algorithm introducing the analysis of permutations of sub-trees, but this would heavily affect performances.
Despite this intrinsic limit, this technique appears to fit very well to our purpose of measuring HTML trees similarity.

It is important to remark that, applying \emph{simple tree matching} to compare simple and quite different trees will produce a more accurate result. 
Despite that, because of the most of modifications in Web pages are usually slight changes, \emph{clustered tree matching} is far and away the best algorithm to be adopted in building automatically adaptable wrappers. 
Moreover, this algorithm makes it possible to establish a custom level of accuracy of the process of matching, defining a similarity threshold required to match two trees.

\section{\uppercase{Adaptable Web Wrappers}}

\noindent Based on the adaptation algorithms described above, a proof-of-concept extension to the Lixto Visual Developer (VD) has been implemented. 
Wrappers are automatically adapted based on given configuration settings, integrity constraints, and triggers. 

Usually, wrapper generation in VD is a hierarchical top-down process, e.g. first, a ``hotel record'' is characterized, and inside the hotel record, entities such as ``rating'' and ``room types''. 
Such entities are referred to as \emph{patterns}. 
To define a pattern, the wrapper designer visually selects an example and together with system suggestions generalizes the rule configuration until the desired instances are matched. 

In this extension, to support the automatic adaptation process during runtime, the wrapper designer further specifies what it means that extraction failed. 
In general, this means wrong or missing data, and with integrity constraints one can give indications how correct results look like.
Typical \emph{integrity constraints} are:

\begin{itemize}
	\item \emph{Occurrence restrictions}: e.g. minimum and/or maximum number of allowed occurrence of a pattern instance, minimum and/or maximum number of child pattern instances;
	\item \emph{Data types}: e.g. all results of a ``price'' pattern need to be of data type integer.
\end{itemize}

\noindent Integrity constraints can be specified with each pattern individually or be based on a data model (in our case, a given XML Schema). 
In case integrity constraints are violated during runtime, the adaptation process for this particular pattern is started. 

During wrapper creation, the application designer provides a number of configuration settings to this process. 
This includes:

\begin{itemize}
	\item Threshold values;
	\item Priorities/order of adaptation algorithms used;
	\item Flags of the chosen algorithm (e.g. using HTML element name as node label, using id/class attributes as node labels, etc.);
	\item Triggers for bottom-up, top-down and process flow adaptation bubbling;
	\item Whether stored tree-grams and XPath statements are updated based on adaptation results to be additionally used as inputs in future adaptation procedures (reflecting and addressing regular slight changes of a Web page over time).
\end{itemize}

\noindent Used algorithms for adaptations rely on two inputs (stored example tree-gram(s), DOM tree of current page) and provide as output sub-trees that are sufficiently similar to the original (example) ones, and in consequence a generated XPath statement that matches the nodes (Fig. \ref{diagram} summarizes the process from design time and execution time perspective). 

Algorithms under consideration include the clustered tree matching discussed above, as well as tree-based variants of the Bigram \cite{Collins1996bigram} and Jaro-Winkler similarity \cite{winkler1999state} (which are of advantage when one assumes that permutations in the tree nodes are likely over time). 
Moreover, for extraction of leaf nodes which exhibit no inherent tree structure, we rely on string similarity metrics.
Finally, triggers in adaptation settings can be used to force adaptation of further fragments of the wrapper:

\begin{itemize}
	\item Top-down: forcing adaptation of all/some descendant patterns (e.g. adapt the ``price'' pattern as well to identify prices within a record if the ``record'' pattern was adapted).
	\item Bottom-up: forcing adaptation of a parent pattern in case adaptation of a particular pattern was not successful. Experimental evaluation pointed out that in such cases it is often the problem that the parent pattern already provides wrong or missing results (even if matched by the integrity constraints) and has to be adapted first.
	\item Process flow: it might happen that particular patterns are no longer detected because the wrapper evaluates on the wrong page. Hence, there is the need to use variations in the deep web navigation processes. A simple approach explored at this time is to use a switch window or back step action to check if the previous window or another tab/pop-up provides the required information.
\end{itemize}

\section{\uppercase{Experimental Results}}
\noindent The best way of measuring reliability of automatically adaptable wrappers is to test their behavior in real world use-cases.
Several common areas of application of Web wrappers have been identified: social networks and bookmarking, retail market and comparison shopping, Web search and information distribution, and, finally, Web communities.
For each of these fields, we designed a test using a representative Website, studying a total of 7 use-cases, defining wrappers applied to 70 Web pages. 
Websites like Facebook, Google News, Ebay, etc. are usually subjected to countless, although often invisible, structural modifications; thus, altering the correct behavior of Web wrappers.
Table \ref{tab-res} summarizes results: each wrapper automatically tries to adapt itself using both the algorithms described in Section 3.
Column referred as \emph{thresh.} means the threshold value of similarity required to match two elements. Columns \emph{tp}, \emph{fp} and \emph{fn} represent true and false positive, and false negative, measures usually adopted to evaluate precision and recall of these kind of tasks.

\begin{table}[!ht]
	\small
	\begin{tabular}{|@{}c@{}  c@{}|@{}c  c  c@{}|@{}c  c  c@{}|}
	
		\cline{3-8}
        \multicolumn{2}{r|}{} & \multicolumn{3}{c|}{Simple T. M.} & \multicolumn{3}{c|}{Clustered T. M.} \\
    \cline{3-8}
        \multicolumn{2}{r|}{} & \multicolumn{3}{c|}{Precision/Recall} & \multicolumn{3}{c|}{Precision/Recall}\\
    \cline{3-8}

		\noalign{\smallskip}    
    \hline
        Scenario & thresh. & tp & fp & fn & tp & fp & fn \\
    \hline
        Delicious & 40\% & 100 & 4 & - & 100 & - & - \\
        Ebay & 85\% & 200 & 12 & - &  196 & -  & 4 \\
        Facebook & 65\% & 240& 72 & - &  240&12 & - \\
				Google news & 90\% & 604 & - & 52 &  644 & - & 12\\
        Google.com & 80\% & 100 & - & 60 &  136 & - & 24 \\
        Kelkoo & 40\% & 60 & 4 & - & 58 & - & 2 \\
        Techcrunch & 85\% &  52 & - & 28 &  80 & - & - \\
    \hline
        Total  & - & 1356 & 92 & 140 & 1454 & 12 & 42\\
    \hline
    \hline
        Recall  & - & \multicolumn{3}{c|}{90.64\%} & \multicolumn{3}{c|}{97.19\%}\\
        Precision  & - & \multicolumn{3}{c|}{93.65\%} & \multicolumn{3}{c|}{99.18\%}\\
        F-Measure  & - & \multicolumn{3}{c|}{92.13\%} & \multicolumn{3}{c|}{98.18\%}\\
		\hline

	\end{tabular}
	
	\caption{Evaluation of the reliability of automatically adaptable wrappers applied to real world scenarios.}
	\label{tab-res}

\end{table}

\noindent Performances obtained using the \emph{simple} and the \emph{clustered} tree matching are, respectively, good and excellent; \emph{clustered tree matching} definitely is a viable solution to automatically adapt wrappers with a high degree of reliability (F-Measure greater than 98\%). 
This system provides also the possibility of improving these results including additional checks on \emph{comparable attributes} (e.g. \emph{id}, \emph{name} or \emph{class}). The role of the required accuracy degree is fundamental; experimental results help to highlight the following considerations: very high values of threshold could result in false negatives (e.g. Google news and Google.com scenarios), while low values could result in false positives (e.g. the Facebook scenario). Our solution exploiting the \emph{clustered tree matching} algorithm, designed by us, helps to reduce wrapper maintenance tasks, keeping in mind that, in cases of deep structural changes, it could be required to manually intervene to fix a specific wrapper, since it is impossible to automatically face all the possible malfunctionings.

\section{\uppercase{Conclusion}}
\noindent In this paper we described several novel ideas, investigating the possibility of designing smart Web wrappers which automatically react to structural modifications of underlying Web pages and adapting themselves to avoid malfunctionings or corrupting extracted data.
After explaining the core algorithms on which this system relies, we shown how to implement this feature in Web wrappers.
Finally, we analyzed performances of this system through a rigorous testing of the behavior of automatically adaptable wrappers in real world use-cases. 

This work opens new scenarios on wrapper adaptation techniques and is liable to several improvements: first of all, avoiding some limitations of the matching algorithms, for example the inability of handling permutations on nodes previously explained, with computationally efficient solutions could be important to improve the robustness of wrappers. 
One limitation of adopted tree matching algorithms is also that they do not work very well if new levels of nodes are added or node levels are removed. 
We already investigated the possibility of adopting different tree similarity algorithms, working better in such cases. 
We could try to ``generalize'' other similarity metrics on strings, such as the n-gram distance and the Jaro-Winkler distance. 
Implementing these two metrics do not require dynamic programming and might be computationally efficient; in particular, variants of the Bigram distance on trees might work well with permutations of groups of nodes and the Jaro-Winkler distance could better reflect missing or added node levels. 
Another idea is investigating the possibility of improving matching criteria including additional information to be compared during the tree match up process (e.g. full path information, all attributes, etc.); then, exploiting logic-based rules (e.g. regular expressions, string edit distance, and so on) to analyze textual properties. 

Finally, the tree-grammar, already exploited to store a light-weight signature of the structure of elements identified by the wrapper, could be extended for  classifying topologies of templates frequently shown by Web pages, in order to define \emph{standard protocols} of automatic adaptation in these particular contexts.
Adaptation in the deep web navigation is a different topic than adaptation on a particular page, but also extremely important for wrapper adaptation. 
Future work will comprise to investigate focused spidering techniques: instead of explicit modeling of a work flow on a Web page (form fill-out, button clicks, etc.) we develop a tree-grammar based approach that decides for a given Web page which template it matches best and executes the data extraction rules defined for this template. 
Navigation steps are carried out implicitly by following all links and DOM events that have been defined as interesting, crawling a site in a focused way to find the relevant information.

Concluding, the system of designing automatically adaptable wrappers described in this paper has been proved to be robust and reliable. 
The \emph{clustered tree matching} algorithm is very extensible and it could be adopted for different tasks, also not strictly related to Web wrappers (e.g. operations that require  matching up trees could exploit this algorithm).

\renewcommand{\baselinestretch}{0.98}
\bibliographystyle{apalike}
{\small
\bibliography{biblio}}
\renewcommand{\baselinestretch}{1}

\end{document}